\documentclass[a4paper]{article}
\usepackage[square,numbers,sort]{natbib}
\usepackage{natbib}
\usepackage[final]{nips_2017}
\usepackage[utf8]{inputenc} 
\usepackage[T1]{fontenc}    
\usepackage{hyperref}       
\usepackage{url}            
\usepackage{booktabs}       
\usepackage{amsfonts}       
\usepackage{nicefrac}       
\usepackage{microtype}      
\usepackage{graphicx}
\usepackage{amsmath}
\usepackage[dvipsnames]{xcolor}
\usepackage{algorithm}
\usepackage{algorithmic}
\usepackage[justification=centering,font=small]{caption}

\usepackage{fancyvrb}

\RecustomVerbatimCommand{\VerbatimInput}{VerbatimInput}%
{fontsize=\footnotesize,
 frame=lines,  
 framesep=2em, 
 rulecolor=\color{Gray},
 label=\fbox{\color{Black}data.txt},
 labelposition=topline,
 commandchars=\|\(\), 
 commentchar=*        
}

\title{Airbnb Price Prediction Using Machine Learning and Sentiment Analysis}

\author{
  Pouya Rezazadeh Kalehbasti \\
  Stanford University\\
  \texttt{pouyar@stanford.edu} \\
  \And
  Liubov Nikolenko\\
  Stanford University\\
  \texttt{liubov@stanford.edu} \\
  \And
  Hoormazd Rezaei \\
  Stanford University\\
  \texttt{hoormazd@stanford.edu} \\
}

\begin{document}

\maketitle
\begin{center}
\end{center}

\setlength{\parindent}{3ex}

\section{Introduction}
Pricing a rental property on Airbnb is a challenging task for the owner as it determines the number of customers for the place. On the other hand, customers have to evaluate an offered price with minimal knowledge of an optimal value for the property. This paper aims to develop a reliable price prediction model using machine learning, deep learning, and natural language processing techniques to aid both the property owners and the customers with price evaluation given minimal available information about the property. Features of the rentals, owner characteristics, and the customer reviews will comprise the predictors, and a range of methods from linear regression to tree-based models, support-vector regression (SVR), K-means Clustering (KMC), and neural networks (NNs) will be used for creating the prediction model.

\section{Related Work}
Parts of the existing literature on property pricing focus on non-shared property purchase or rental price predictions. Previously, Yu and Wu \cite{yureal} tried to implement a real estate price prediction using feature importance analysis along with linear regression, SVR, and Random Forest regression. They also attempted to classify the prices into 7 classes using Naive Bayes, Logistic Regression, SVC and Random Forest. They declared a best RMSE of 0.53 for their SVR model and a classification accuracy of 69\% for their SVC model with PCA. In another paper, Ma et al. \cite{ma2018estimating} have applied Linear Regression, Regression Tree, Random Forest Regression and Gradient Boosting Regression Trees to analyzing warehouse rental prices in Beijing. They concluded that the tree regression model was the best-performing model with an RMSE of 1.05 CNY/$m^2$-day

Another class of studies, which are more pertinent to this work, inspect the hotels and sharing economy rental prices. In a recent work, Wang and Nicolau \cite{wang2017price} have studied price determinants of sharing economy by analyzing Airbnb listings using ordinary least squares and quantile regression analysis. In a similar study, Masiero et al. \cite{masiero2015demand} use quantile regression model to analyze the relation between travel traits and holiday homes as well as hotel prices. In a simpler work, Yang et al. \cite{yang2016market} applied linear regression to study the relationship between market accessibility and hotel prices in Caribbean. They also included the user ratings and hotel classes as contributing factors in their study. Li et al. \cite{li2016reasonable} also studied a clustering method called Multi-Scale Affinity Propagation and applied Linear Regression to the obtained clusters in an effort to create a price prediction model for Airbnb in different cities. They took the distance of the property to the city landmarks as the clustering feature.

This research has tried to improve and add to the experimented methods from the literature by focusing on a variety of feature selection techniques, implementing Neural Networks, and leveraging the customer reviews through sentiment analysis. The last two contributions are novel undertakings in rental price prediction as they were not observed in the existing body of literature.

\section{Dataset}
\begin{figure}[h]
    \centering
    \includegraphics[scale=0.4]{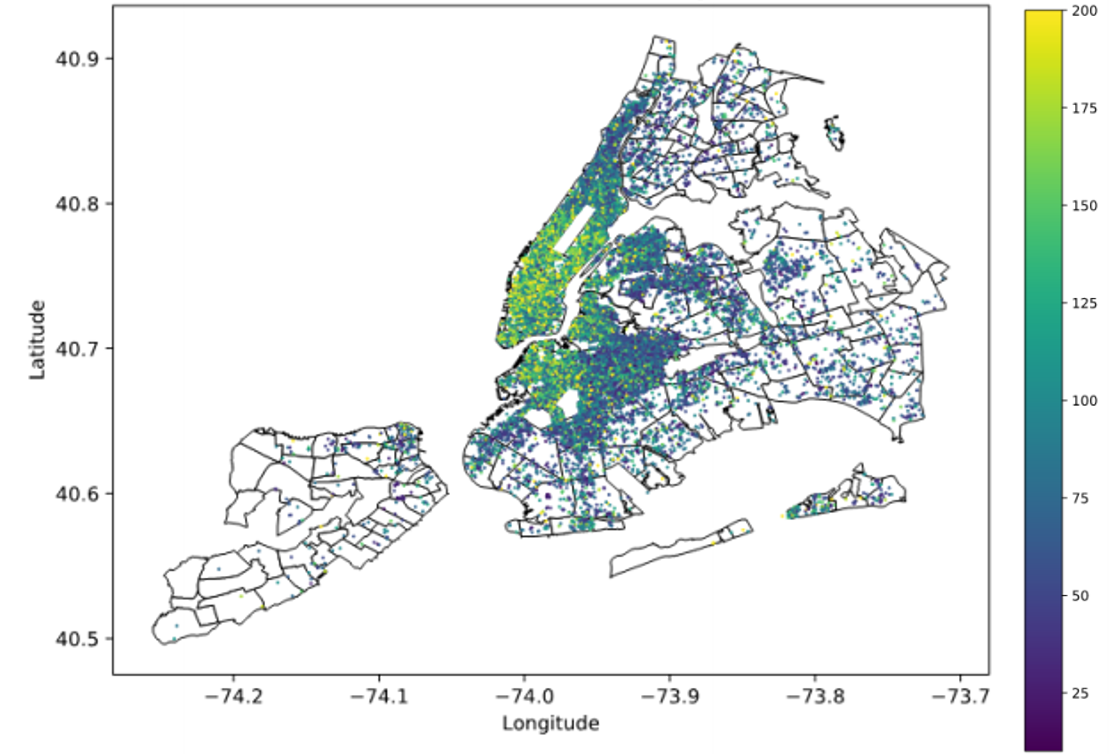}
    \caption{Geographic spread of price labels (with filtered outliers)}
    \label{fig:heatmap}
\end{figure}

The public Airbnb dataset for New York City \cite{AirbnbData} was used as the main data source for this study. The dataset included 50,221 entries, each with 96 features. Figure \ref{fig:heatmap} shows the geographic distribution of the listing prices in this dataset.\\
\\
 For the initial prepossessing, the authors inspected each feature of the dataset to (i) remove features with frequent and irreparable missing fields or set the missing values to zero where appropriate, (ii) convert some features into floats (e.g. by removing the dollar sign in prices), (iii) change boolean features to binaries, (iv) remove irrelevant or uninformative features, e.g. host picture url, constant-valued fields or duplicate features, and (v) convert the 10 categorical features in the final set, e.g. `neighborhood name' and `cancellation policy,' into "one-hot vectors." In addition, the features were normalized and the labels were converted into logarithm of the prices to mitigate the impact of the outliers in the dataset. The data was split into three sets; namely, train set (comprising 90\% of the original data), validation set, and test set (both comprising 5\% of original data). Since the dataset was relatively large, 10\% of the data was deemed sufficient for the accumulated testing and validation sets. The following explains the sentiment analysis conducted on the reviews and the steps taken for selecting the most important features among the available set of features.
 
\subsection{Sentiment Analysis on the Reviews}
Given the importance of customer reviews on the pricing of an Airbnb listing, and in order to increase the accuracy of the predictive model, the reviews for each listing were analyzed using TextBlob \cite{loria2014textblob} sentiment analysis library and the results were added to the set of features. This method assigns a score between -1 (very negative sentiment) and 1 (very positive sentiment) to each analyzed text. For every listed property, each reviews was analyzed using this method and the scores were averaged across all the reviews associated with that listing. The final scores for each listing was included as a new feature in the model.

\subsection{Feature Selection}
After data preprocessing, the feature vector contained 764 elements. Feeding this excessive set of features to the models resulted in a high variance of error. Consequently, several feature selection techniques were used to find the features with the most predictive values to both reduce the model variances and reduce the computation time. Based on prior experience of the authors with housing price estimation, the first tried method was manual selection of features to create a baseline for evaluating the other feature selection processes.

The second selection method was tuning the coefficient of linear regression model with Lasso Regularization trained on the train split. Based on this analysis, the model with the best performance over validation split was selected. The resulting set consisted of 78 features with non-zero values, i.e. 90\% less than the number of original features.

Finally, lowest p-values of regular linear regression model trained on train split were used to choose the third set of features. An upper limit of 100 features was imposed on the selection procedure. The final set was comprised of 22 features for which linear regression model performed the best on the validation split. As an example to demonstrate the results of the feature selection techniques, Appendix \ref{AppendixA} lists the set of features resulting from this p-value analysis.

\begin{figure}[h]
    \centering
    \includegraphics[scale=0.5]{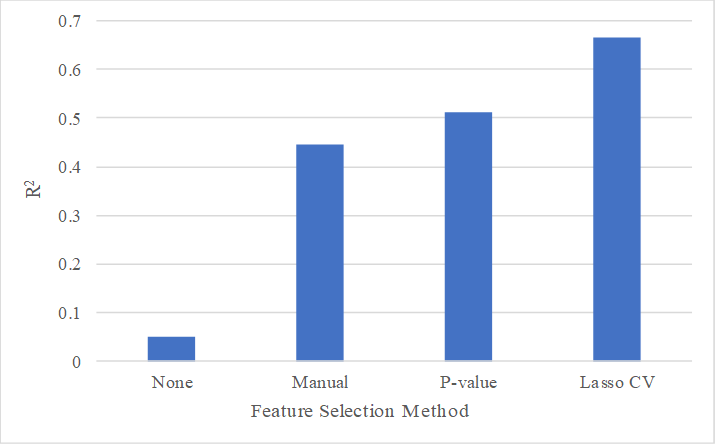}
    \caption{Best feasible $R^2$ scores with each selection methods}
    \label{fig:feature_select}
\end{figure}

The performance of manually selected features as well as p-value and Lasso feature selection schemes were compared using the $R^2$ score of the linear regression models trained on the validation set. All models outperformed the baseline model, which used the whole feature set, and the second method, Lasso regularization, yielded the highest $R^2$ score. Figure \ref{fig:feature_select} shows the best $R^2$ scores obtained using the set of features identified with each feature selection method.

\section{Methods}
Linear Regression using the entire set of features as model inputs was taken as the baseline model for evaluating the performance of the other methods. After selecting a set of features using Lasso feature selection, several machine learning models were considered in order to find the optimal one. All of the models except neural networks were implemented using Scikit-learn library \cite{sklearn}. The neural network model was implemented with the help of Keras library \cite{keras}. The implemented models are introduced in what follows.

\subsection{Ridge Regression}
Linear Regression with $L_2$ regularization adds a penalizing term to the squared error cost function in order to help the algorithm converge for linearly separable data and reduce overfitting. Therefore, Ridge Regression minimizes $J(\theta) = ||y - X\theta||^2_2 + \alpha||\theta||_2^2$ with respect to $\theta$, where $X$ is a design matrix and $\alpha$ is a hyperparameter. Since the baseline models were observed to have high variance, Ridge Regression seemed to be an appropriate choice to solve the issue.
\subsection{K-means Clustering with Ridge Regression}
In order to capture the non-linearity of the data, the training examples were split into different clusters using k-means clustering on the features and the Ridge Regression was run on each of the individual clusters. The data clusters were identified using the following algorithm:
\begin{algorithm}
\caption{K-means Clustering}
\begin{algorithmic}
\STATE Initialize cluster centroids $\mu_i, ..., \mu_k$ randomly
\REPEAT 

\STATE Assgin each point to a cluster: $c^{(i)} = \arg \min_j ||x^{(i)} - \mu_j||_2^2$
\STATE For each centroid: $\mu_j = \frac{\sum_{i = 1}^m 1\{c^{(i)} = j\}x^{(i)}}{\sum_{i = 1}^m 1\{c^{(i)} = j\}}$

\STATE Calculate the loss function for the assignments and check for convergence:
\STATE $J(c, \mu) = \sum_{i = 1}^m ||x^{(i)} - \mu_{c^{(i)}}||_2^2$
\UNTIL{convergence}
\end{algorithmic}
\end{algorithm}

\subsection{Support Vector Regression}
In order to model the non-linear relationship between the covariates, the authors employed support vector regression with RBF kernel to identify a linear boundary in a high-dimensional feature space. Using the implementation based on Chang and Lin \cite{chang2011libsvm}, the algorithm provides a solution for the following optimization problem:
\begin{align}
\min_{w, b, \xi, \xi^*} \frac{1}{2}||w||^2 + C \sum_{i = 1}^m \xi_i +  C \sum_{i = 1}^m \xi_i^*, \text{subject to}\\
w^T\phi(x^{(i)}) + b - y^{(i)} \leq \epsilon + \xi_i,\\
y^{(i)} - w^T\phi(x^{(i)}) - b  \leq \epsilon + \xi_i^*,\\
\xi_i, \xi_i^* \geq 0, i = 1, ..., m
\end{align}
where $C > 0, \epsilon > 0$ are given parameters. This problem can be converted into a dual problem that does not involve $\phi(x)$, but involves $K(x, z) = \phi(x)\phi(z)$ instead. Since we are using RBF kernel, $K(x, z)$ was taken as
\begin{align}
    K(x, z)=\exp\left(\frac{||x - z||^2}{2 \sigma^2}\right)
\end{align}

\subsection{Neural Network}
Neural network was used to build a model that combined the input features into high level predictors. The architecture of the optimized network had 3 fully-connected layers: 20 neurons in the first hidden layer with relu activation function, 5 neurons in the second hidden layer with relu activation function, and 1 output neuron with a linear activation function.

\subsection{Gradient Boosting Tree Ensemble}
Since the relationship between the feature vector and price is non-linear, regression tree seemed like a proper model for this problem. Regression trees split the data points into regions according to the following formula 
\begin{align}
\max_{j, t} L(R_p) - (L(R_1) - L(R_2))
\end{align}
where $j$ is the feature the dataset is split on, $t$ is the threshold of the split, $R_p$ is the parent region and $R_1$ and $R_2$ are the child regions. Squared error is used as the loss function. 

Since standalone regression trees have low predictive accuracies individually, gradient boost tree ensemble was used to increase the models' performance. The idea behind a gradient boost is to improve on a previous iteration of the model by correcting its predictions using another model based on the negative gradient of the loss. The algorithm for the gradient boosting is the following \cite{johansson}:
\begin{algorithm}
\caption{Gradient Boosting}
\begin{algorithmic}
\STATE Initialize $F_0$ to be a constant model
\FOR{m = 1,..., number of iterations}
\FORALL{training examples $(x^{(i)}, y^{(i)})$} 

\STATE Squared error $R(y^{(i)}, F_{m - 1}(x^{(i)})) = -\frac{\partial \text{Loss}}{\partial F_{m - 1}(x^{(i)})} = y^{(i)} - F_{m - 1}(x^{(i)})$
\ENDFOR
\STATE Train regression model $h_m$ on $(x^{(i)}, R(y^{(i)}, F_{m - 1}(x^{(i)})))$, for all training examples
\STATE $F_m(x) = F_{m - 1}(x) + \alpha h_m(x)$, where $\alpha$ is the learning rate

\ENDFOR
\RETURN $F_m$
\end{algorithmic}
\end{algorithm}

\section{Experiments and Discussion}
Mean absolute error (MAE), mean squared error (MSE) and $R^2$ score were used to evaluate the trained models. Training (39,980 examples) and validation (4,998 examples) splits were used to choose the best-performing models within each category. The test set, containing 4,998 examples,  was used to provide an unbiased estimate of error, with the final models trained on both train and validation splits. Table \ref{table:models} contains the performance metrics for the final models\footnote{Optimized models can be found at \url{github.com/PouyaREZ/AirBnbPricePrediction.git}}; namely, linear regression, Ridge regression, Gradient Boosting, K-Means Clustering with Ridge Regression, SVR, and Neural Network.
\begin{center}
    \begin{table}
         \caption{Performance metrics of the trained models}
         \label{table:models}
         \begin{tabular} {||c c c c c c c||} 
         \hline
         Model Name &  & train split &  &  & test split &  \\ [0.4ex]
         & MAE & MSE & $R^2$ Score & MAE & MSE & $R^2$ Score \\
         \hline\hline
         Linear Reg. (Baseline) & 0.2744 &   0.1480 &  0.690 &  96895.82 & 2.4E13 & -5.1E13 \\ 
         \hline
         Ridge Reg. & 0.2813 & 0.15461 &  0.6765 & 0.2936 & 0.1613 & 0.6601 \\
         \hline
         Gradient Boost & 0.2492 & 0.1376 & 0.7121 & 0.3282 & 0.1963 & 0.5864 \\
         \hline
         K-means + Ridge Reg. & 0.2717 &  0.1438 & 0.6992 & 0.2850 & 0.1543 & 0.6748 \\
         \hline
         SVR & 0.2132 & 0.1067 &  0.7768 & 0.2761 & 0.1471 & 0.6901 \\
         \hline
         Neural Net & 0.2602 & 0.1316 & 0.7246 & 0.2881 & 0.1570 & 0.6692 \\
         \hline
        \end{tabular}
    \end{table}
\end{center}
The outlined models had relatively similar $R^2$ scores which implies that Lasso feature importance analysis had made the most impact on improving the performance of the models by reducing the variance. Even after the feature selection, the resulting input vector was relatively large leaving room for model overfitting. This explains why Gradient Boost - a tree-based model prone to high variance - performed worse than the rest of the models despite it not performing the worst on the train set.

\begin{figure}[h]
    \centering
    \includegraphics[scale=0.4]{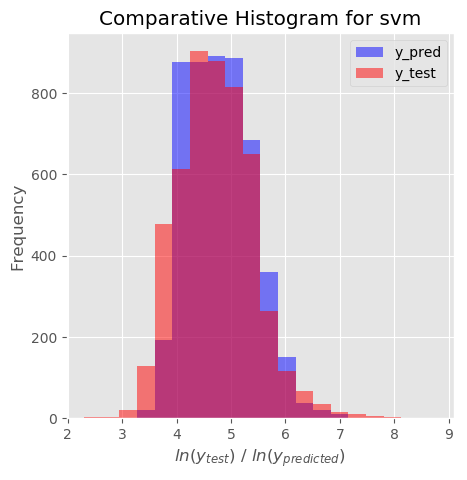}
    \includegraphics[scale=0.4]{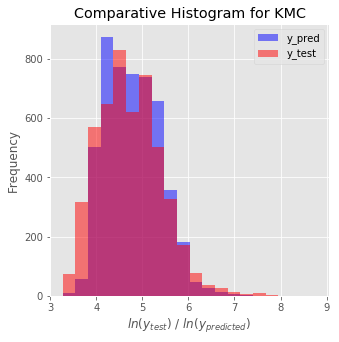}
    \includegraphics[scale=0.4]{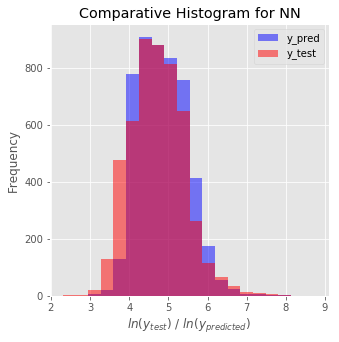}
    \caption{Comparative histograms of predicted and actual prices for the top 3 models: SVR, KMC, and NN}
    \label{fig:svm_hist}
\end{figure}

Despite expanding the number of features in the feature vector, SVR with RBF kernel turned out to be the best performing model with the least MAE and MSE and the highest $R^2$ score on both train and test sets (figure \ref{fig:svm_hist}). RBF feature mapping was able to better model the prices of the apartments which have a non-linear relationship with the apartment features. Since regularization is taken into account in the SVR optimization problem, parameter tuning ensured that the model was not overfitting. 
\\
Ridge regression, neural network, K-means + Ridge regression models had similar $R^2$ scores even though the last two models are more complex than Ridge regression. The architecture complexity of neural network was limited by the insufficient number of training examples for having too many unknown weights. K-means clustering model faced a similar issue: since the frequency of some prices was greatly exceeding the frequency of others, some clusters received too few training examples and drove down the overall model performance.   
\section{Conclusions and Future Work}
This paper attempts to come up with the best-performing model for predicting the Airbnb prices based on a limited set of features including property specifications, owner information, and customer reviews on the listings. Machine learning techniques including linear regression, tree-based models, SVR, and neural networks along with feature importance analyses are employed to achieve the best results in terms of Mean Squared Error, Mean Absolute Error, and $R^2$ score. The initial experimentation with the baseline model proved that the abundance of features leads to high variance and weak performance of the model on the validation set compared to the train set. Lasso-based feature importance analysis reduced the variance and using advanced models such as SVR and neural networks resulted in higher $R^2$ score for both the validation and test sets. Among the models tested, Support Vector Regression (SVR) performed the best and produced an $R^2$ score of 69\% and a MSE of 0.147 (defined on ln(price)) on the test set. This level of accuracy is a promising outcome given the heterogeneity of the dataset and the involved hidden factors and interactive terms, including the personal characteristics of the owners, which were impossible to consider.

The future works on this study can include (i) studying other feature selection schemes such as Random Forest feature importance, (ii) further experimentation with neural net architectures, and (iii) getting more training examples from other hospitality services such as VRBO to boost the performance of K-means clustering with Ridge Regression model in particular.

\newpage
\appendix
\section{Appendix A} \label{AppendixA}
List of features selected using p-value importance method (the last 6 feature names are those of one-hot vectors):\\
\indent `longitude', `accommodates', `bathrooms', `bedrooms', `beds', `security\_deposit', `cleaning\_fee', `guests\_included', `Cable\_TV', `Dryer', `Washer', `Family/kid\_friendly', `Gym', `Elevator', `Entire home/apt', `Private room', `Brooklyn', `Manhattan', `Brooklyn.1', `New York', `Chelsea', `Midtown'

\bibliographystyle{ieeetr}
\bibliography{references}

\end{document}